\documentclass{article}
\usepackage{ijcai21}

\usepackage{times}
\usepackage{subfigure}

\usepackage{soul}
\usepackage{url}
\usepackage[hidelinks]{hyperref}
\usepackage[utf8]{inputenc}
\usepackage[small]{caption}
\usepackage{graphicx}
\usepackage{amsmath}
\usepackage{cases}
\usepackage{amsthm}
\usepackage{booktabs}
\usepackage{algorithm}
\usepackage{algorithmic}
\usepackage{stfloats}
\usepackage{xspace}
\usepackage{xcolor}
\newcommand{\algorithmname}{Algorithm}
\newcommand{\equationname}{Eq.}
\newcommand{\method}{{FedHealth 2}\xspace}

\title{\method: Weighted Federated Transfer Learning via Batch Normalization for Personalized Healthcare}

\author{
Yiqiang Chen$^{1,2}$\footnote{Corresponding Author}\and
Wang Lu$^{1,2}$\and
Jindong Wang$^3$\And
Xin Qin$^{1,2}$\\
\affiliations
$^1$Institute of Computing Technology, Chinese Academy of Sciences, Beijing, China\\
$^2$University of Chinese Academy of Sciences, Beijing, China\\
$^3$Microsoft Research Asia, Beijing, China\\
\emails
\{yqchen,luwang,qinxin18b\}@ict.ac.cn, jindong.wang@microsoft.com
}

\begin{document}

\maketitle

\begin{abstract}

The success of machine learning applications often needs a large quantity of data. Recently, federated learning (FL) is attracting increasing attention due to the demand for data privacy and security, especially in the medical field. However, the performance of existing FL approaches often deteriorate when there exist domain shifts among clients, and few previous works focus on personalization in healthcare. In this article, we propose \method, an extension of FedHealth~\cite{chen2020fedhealth} to tackle domain shifts and get personalized models for local clients. \method obtains the client similarities via a pretrained model, and then it averages all weighted models with preserving local batch normalization. Wearable activity recognition and COVID-19 auxiliary diagnosis experiments have evaluated that \method can achieve better accuracy (\textbf{10}\%+ improvement for activity recognition) and personalized healthcare without compromising privacy and security.

\end{abstract}

\section{Introduction}
For the past few years, machine learning technology has been widely used in real life. Machine learning technology leverages the labeled data to learn the patterns and trends of observed objects. With the help of perception technology, well-trained machine learning models liberate people from heavy and repetitive tasks, especially in the healthcare field. A successful machine learning healthcare application often requires sufficient user data to dig into the essence of health status. However, with the increasing awareness of privacy and security of the people and organizations, China, the European Union, and some other governments or organizations enforce the protection of user data via different regulations~\cite{inkster2018china,voigt2017eu}. In this situation, federated learning~\cite{yang2019federated} came into being to protect data privacy and security.

\begin{figure}[htbp!]
	\centering
	\includegraphics[width=0.45\textwidth]{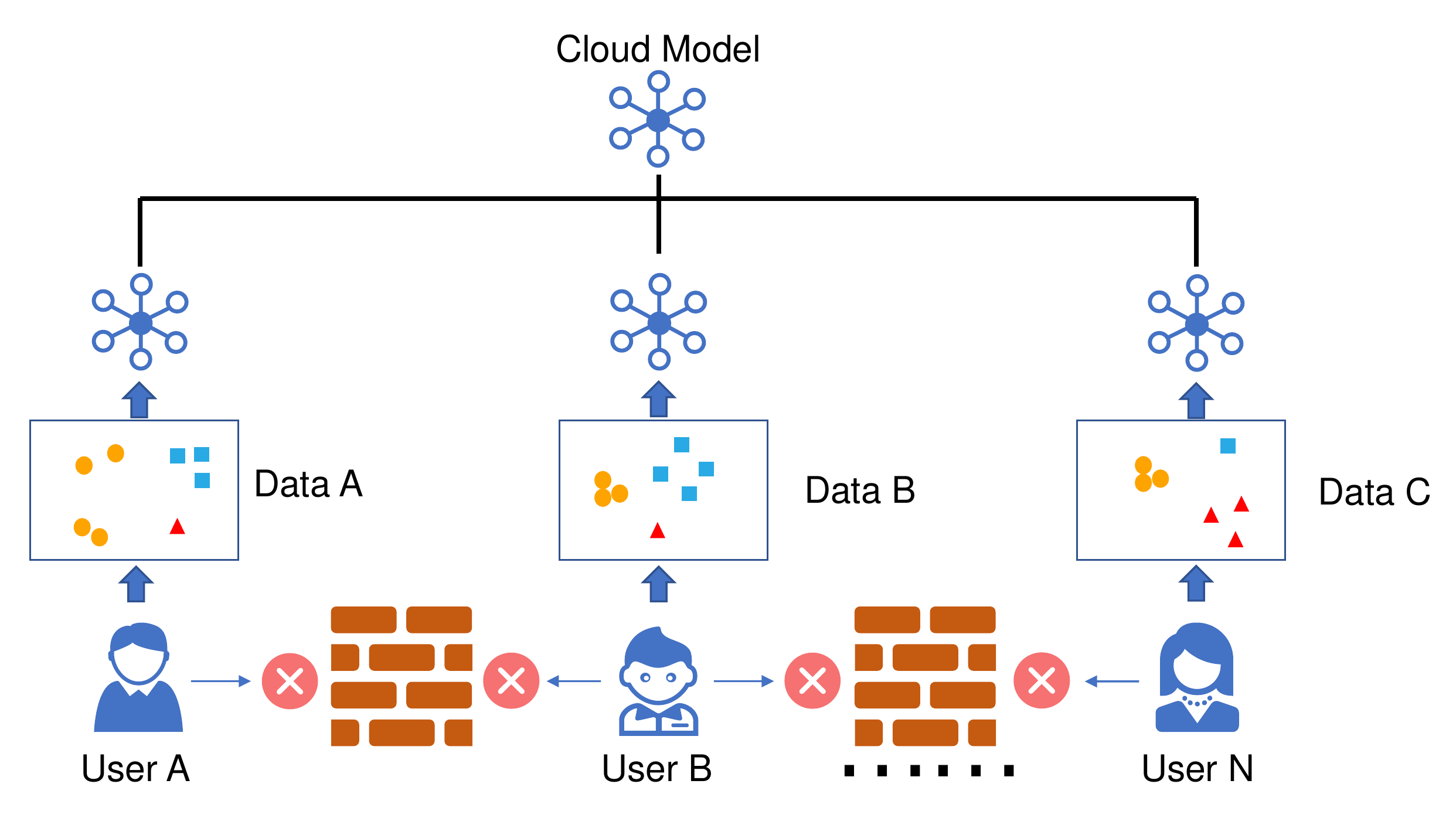}
	\caption{Data islanding and personalization in healthcare.}
	\label{fig:fedavg}
\end{figure}

The most classic algorithm in FL is FedAvg~\cite{mcmahan2017communication}, which directly averages the parameters of models coming from all participating clients without transporting local data. Although FedAvg is simple, it has demonstrated superior performance in many situations. However, FedAvg is not designed particularly for non-iid data, which may make it suffer from performance degradation or even diverge when it is deployed over non-iid samples~\cite{li2019convergence}. As shown in \figurename~\ref{fig:fedavg}, different clients (users) cannot exchange their own data straightforward, and they exchange their information via model parameters with the help of the cloud. Various distributions of the client data make it hard to achieve good performance when directly employing FedAvg. 
The previous work, FedHealth~\cite{chen2020fedhealth}, is the first federated transfer learning framework for wearable healthcare. Although it achieves great success in federated learning for healthcare and solves the problems of data islanding and personalization, it requires a large number of data shared across different clients, which is unrealistic in many situations.

In this article, we propose \method, a weighted federated transfer learning algorithm via batch normalization for personalized healthcare. \method can solve both data islanding and personalization problems without sharing common data. Specifically, \method gets the similarities among clients with the help of a pretrained model. The similarities are determined by the distances of the data distributions, and the distances can be calculated via the statistical values of the layers' outputs of the pretrained network. After obtaining the similarities, the cloud averages the weighted models' parameters in a personalized way and generates a unique model for each client. Each client preserves its own batch normalization and updates the model with a momentum method. Obviously, \method can not only cope with feature shift non-iid~\cite{li2021fedbn} via keeping local batch normalization but also cope with some other shifts, such as label shifts by considering clients' similarities. In addition, all parameter exchange processes are encrypted to ensure that data is not leaked. \method is extensible and can be deployed to many healthcare applications with different expansion methods.

Our contributions are as follows.
\begin{enumerate}
    \item We propose \method, a weighted federated transfer learning algorithm via batch normalization for healthcare, which can aggregate the information from different clients without compromising privacy security, and achieve personalized models for clients through weighting models and preserving local batch normalization.
    \item We show the excellent performance achieved by \method in healthcare applications. Experiments show that \method dramatically improves the recognition accuracy of the local clients' models. At the same time, it reduces the number of rounds and speeds up the convergence to some extent.
    \item \method is extensible and can be employed in many healthcare applications. Diverse extensions allow it can work in many circumstances. Even if there does not exist a pretrained model, \method can get similarities via FedBN~\cite{li2021fedbn} with few rounds.
\end{enumerate}

\section{Related Work}
\subsection{Machine Learning and Healthcare}
With the rapid development of the technology of perception and computing, people can make use of machine learning to help doctors diagnose, assist doctors in the operation, etc. And with the help of perception technology, machine learning even can carry out disease warnings via daily behavior supervision. For instance, certain activities in daily life reflect early signals of some cognitive disease. Through daily observation of gait changes and finger flexibility, the machine can tell people whether they are suffering from Parkinson~\cite{chen2017pdassist}.

Unfortunately, a successful healthcare application needs a large amount of labeled data of users. However, in real applications, data are often separate and few people are willing to disclose their private data. In addition, an increasing number of regulations, such as~\cite{inkster2018china,voigt2017eu}, hold back the leakages of data.

\subsection{Federated Learning}
Federated learning is a usual way to combine each client's information while protecting data privacy and security. It was first proposed by Google~\cite{mcmahan2017communication}, where they proposed FedAvg to train machine learning models via aggregating distributed mobile phones' information without exchanging data. The key idea is to replace direct data exchanges with model parameter-related exchanges. FedAvg is able to resolve the data islanding problems.

Though FedAvg works well in many situations, it still suffers from performance degradation or even diverges when meeting non-iid issues. Some works try to tackle these problems. FedProx~\cite{li2018federated} tackled the heterogeneity by allowing partial information aggregation and adding a proximal term to FedAvg. \cite{yeganeh2020inverse} aggregated the models of the clients with weights computed via $L_1$ distance among client models' parameters. These works focus on a common model shared by all clients while some other works try to obtain a unique model for each client. \cite{arivazhagan2019federated} exchanged base layers' information and preserved personalization layer to combat the ill-effects of heterogeneity. \cite{dinh2020personalized} utilized Moreau envelopes as clients’ regularized loss function and decoupled personalized model optimization from the global model learning in a bi-level problem stylized for personalized FL. \cite{yu2020salvaging} evaluated three techniques for local adaptation of federated models: fine-tuning, multi-task learning, and knowledge distillation. Two works most relevant to our method are FedHealth~\cite{chen2020fedhealth} and FedBN~\cite{li2021fedbn}. FedHealth proposed a federated transfer learning framework which needs some sharing data in healthcare while FedBN used local batch normalization to alleviate the feature shift before averaging models. Although there are already some works to cope with non-iid issues, few works pay attention to feature shift non-iid and other shifts at the same time and getting individual model for each client in healthcare.

\subsection{Batch Normalization}
Batch Normalization (BN)~\cite{ioffe2015batch} has been an indispensable component of deep learning since it was created. Batch Normalization improves the performance of the model and has a natural advantage in dealing with domain shifts. Nowadays, researchers have explored many effects of BN, especially in transfer learning~\cite{segu2020batch}. FedBN~\cite{li2021fedbn} is one of few applications of BN in the field of FL field. However, FedBN does still not make full use of BN properties, and it does not consider the similarities among the clients.

\section{Method}
\subsection{Problem Formulation}
In a FL problem, there are $N$ different clients (organizations or users), denoted as $\{ C_1, C_2, \cdots, C_N \}$ and each client has its own dataset, i.e. $\{ \mathcal{D}_1, \mathcal{D}_2, \cdots, \mathcal{D}_N \}$. Each dataset, $\mathcal{D}_i = \{ (\mathbf{x}_{i,j}, y_{i,j}) \}_{j=1}^{n_i}$, contains two parts, i.e. a train dataset $\mathcal{D}_i^{train} = \{ (\mathbf{x}_{i,j}^{train}, y_{i,j}^{train}) \}_{j=1}^{n_i^{train}}$ and a test dataset $\mathcal{D}_i^{test} = \{ (\mathbf{x}_{i,j}^{test}, y_{i,j}^{test}) \}_{j=1}^{n_i^{test}}$. Obviously, we have $n_i = n_i^{train} + n_i^{test}$ and $\mathcal{D}_i = \mathcal{D}_i^{train} \cup \mathcal{D}_i^{test}$. All of the datasets have different distributions, i.e. $P(\mathcal{D}_i) \neq P(\mathcal{D}_j)$. Each client has its own model denoted as $\{ f_i\}_{i=1}^N$. Our goal is to combine information of all clients to learn a good model $f_i$ for each client on its local dataset $\mathcal{D}_i$ without private data leakage:
\begin{equation}
    \min_{\{f_k\}_{k=1}^N} \frac{1}{N} \sum_{i=1}^N \frac{1}{n_{i}^{test}} \sum_{j=1}^{n_i^{test}} \ell(f_i(\mathbf{x}_{i,j}^{test}), y_{i,j}^{test}),
    \label{eqa:goal}
\end{equation}
where $\ell$ is a loss function.

\subsection{Overview of \method}
\begin{figure*}[t!]
	\centering
	\includegraphics[width=0.8\textwidth]{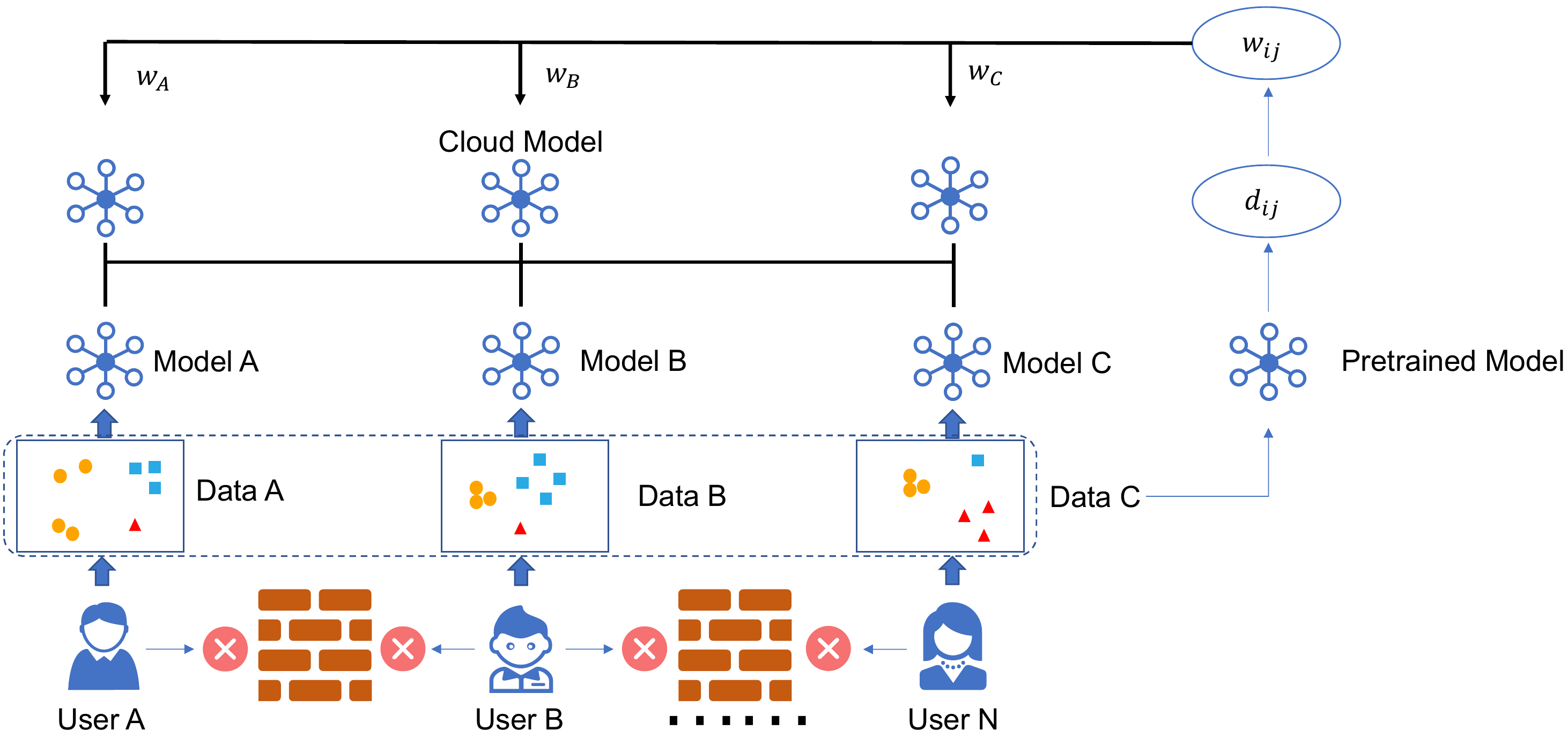}
	\caption{The overall structure of the \method method for federated learning.}
	\label{fig:wfedbn}
\end{figure*}

\method aims to achieve accurate personal healthcare through weighted federated transfer learning via local batch normalization without compromising data privacy and security. \figurename~\ref{fig:wfedbn} gives an overview of the structure. Without loss of generality, we assume there are three clients, which can be extended to the more general case. The structure mainly contains five parts. Firstly, the server distributes the pretrained model to each client. Secondly, each client computes statistics of the outputs of specific layers according to local data. Thirdly, the server obtains clients' similarities represented by the weight matrix $\mathbf{W}$ which will be used to guide aggregation. Fourthly, each client updates its own model with the local train data and pushes their models to the cloud. Finally, the server aggregates models and obtains $N$ models delivered to $N$ clients respectively. All processes do not involve the direct transmission of data, so \method avoids the leakage of data privacy to a certain extent.

The keys of \method are obtaining $\mathbf{W}$ and aggregating the models. The way of computing $\mathbf{W}$ will be introduced in the following, and we first describe how to aggregate the models after obtaining $\mathbf{W}$.

\begin{figure}[htbp!]
	\centering
	\includegraphics[width=0.5\textwidth]{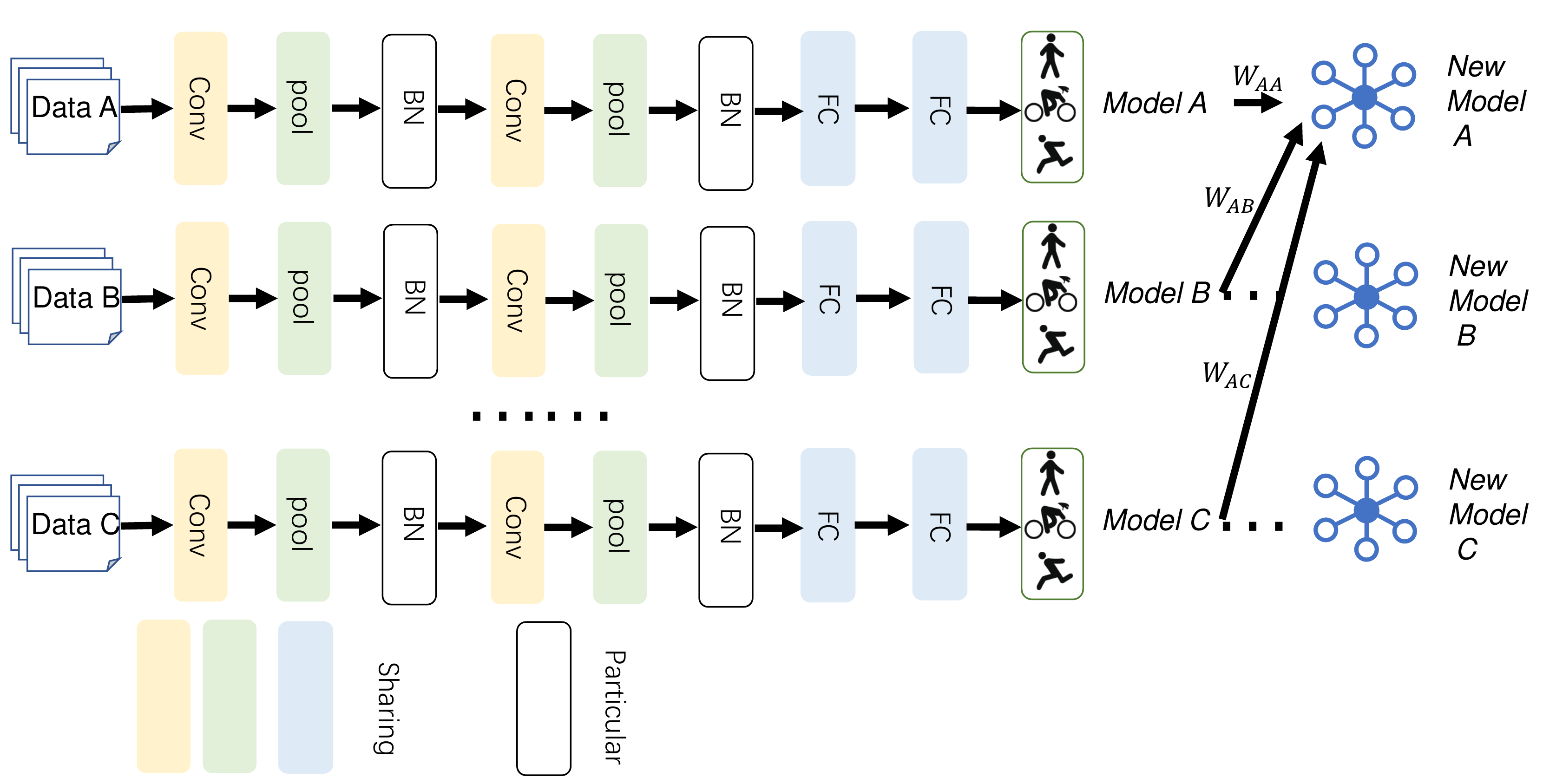}
	\caption{The concrete process of the \method.}
	\label{fig:fedbn}
\end{figure}

We denote the parameters of each model $f_i$ as $\boldsymbol{\theta}_i = \boldsymbol{\phi}_i \cup \boldsymbol{\psi}_i$. $\boldsymbol{\phi}_i$ corresponds to the parameters of BN layers while $\boldsymbol{\psi}_i$ corresponds to the parameters of the other layers. $\mathbf{W}$ is an $N \times N$ matrix, which describes the similarities among the clients. $w_{ij}$ demonstrates the similarity between client $i$ and client $j$. The bigger $w_{ij}$ is, the more similar the two clients are. We have $w_{ii} = \lambda$, where $\lambda$ is a hyper-parameter, and $\sum_{j=1}^N w_{ij} = 1$. 

\figurename~\ref{fig:fedbn} demonstrates the process of aggregating the models. As shown in \figurename~\ref{fig:fedbn}, $\boldsymbol{\phi}_i$ is fixed while $\boldsymbol{\psi}_i$ is computed according to $\mathbf{w}_{i}$, where $\mathbf{w}_{i}$ means the $i$th rows of $\mathbf{W}$, and $\boldsymbol{\psi}$, where $\boldsymbol{\psi} = \{ \boldsymbol{\psi}_i \}_{i=1}^N$. Let $\boldsymbol{\theta}_i^t = \boldsymbol{\phi}_i^t \cup \boldsymbol{\psi}_i^t $ represents the parameters of the model form client $i$ in the round $t$, then we have
\begin{equation}
    \begin{cases}
    \boldsymbol{\phi}_i^{t+1}  = & \boldsymbol{\phi}_i^t\\
    \boldsymbol{\psi}_i^{t+1}  = & \sum_{j=1}^N w_{ij}\boldsymbol{\psi}_j^t.
    \end{cases}
    \label{eqa:updaterule}
\end{equation}

The overall process of \method is described in \algorithmname~\ref{alg:algorithm}.
In next sections, we will introduce how to compute the weight matrix $\mathbf{W}$ with or without a pretrained model.

\begin{algorithm}[tb]
\caption{\method}
\label{alg:algorithm}
\textbf{Input}: A pretrained model $f$, $N$ clients' datasets $\{\mathcal{D}_i\}_{i=1}^N$, $\lambda$\\
\textbf{Output}: Client models $\{f_i\}_{i=1}^N$
\begin{algorithmic}[1] 
\STATE Distribute $f$ to each client
\STATE Each client computes its statistics $(\boldsymbol{\mu}_i, \boldsymbol{\sigma}_i)$, where $\boldsymbol{\mu}_i$ represents the mean values while $\boldsymbol{\sigma}_i$ represents the covariance matrixs. Push $(\boldsymbol{\mu}_i, \boldsymbol{\sigma}_i)$ to the cloud
\STATE Compute $\mathbf{W}$ according to the statistics
\STATE Update the client model with local data. Push $\{ \boldsymbol{\theta}_i^t \}_{i=1}^N$ to the cloud
\STATE Update $\{ \boldsymbol{\theta}_i^{t} \}_{i=1}^N$ according to \equationname~\ref{eqa:updaterule} and distribute $\{ \boldsymbol{\theta}_i^{t+1} \}_{i=1}^N$ to the corresponding clients \STATE Repeat steps $4 \sim 5$ util convergence or maximum round
\end{algorithmic}
\end{algorithm}
\subsection{Evaluate Weights With a Pretrained Model}
In this section, we will evaluate the weights with a pretrained model $f$ and propose two ways to compute the weights. 

We denote with a $l \in \{1, 2, \cdots, L \}$ in superscript notations the different batch normalization layers in the model. And $\mathbf{z}^{i,l}$ represents the input of $l$th batch normalization layer in the $i$th client. The input of the classify layer in the $i$th client is denoted as $\mathbf{z}^i$ which represents the domain features. We assume $\mathbf{z}^{i,l}$ is a matrix, $\mathbf{z}^{i,l}_{c_{i,l}\times s_{i,l}}$ where $c_{i,l}$ corresponds to the channel number while $s_{i,l}$ is the product of the other dimensions. Similarly, $\mathbf{z}^i = \mathbf{z}^i_{c_i \times s_i}$. We feed $\mathcal{D}_i$ into $f$, and we can obtain $\mathbf{z}^{i,l}_{c_{i,l}\times s_{i,l}}$. Obviously, $s_{i,l} = e \times n_i$ where $e$ is an integer. Now, we try to compute statistics on the channels, and we treat $\mathbf{z}^{i,l}$ as a Gaussian distribution. For the $l$th layer of the $i$th client, it is easy to obtain its distribution parameters, $\mathcal{N}(\boldsymbol{\mu}^{i,l}, \boldsymbol{\sigma}^{i,l})$. And the $i$th client's statitics are denoted as
\begin{equation}
    (\boldsymbol{\mu}_i,\boldsymbol{\sigma}_i) = [ (\boldsymbol{\mu}^{i,1},\boldsymbol{\sigma}^{i,1}), (\boldsymbol{\mu}^{i,2},\boldsymbol{\sigma}^{i,2}), \cdots, 
(\boldsymbol{\mu}^{i,L},\boldsymbol{\sigma}^{i,L})].
\label{eqa:zl}
\end{equation}

Now we can calculate the distance between two clients. We usually utilize Wasserstein distance to calculate the distance between two Gaussian distributions. According to \cite{peyre2019computational}, \begin{equation}
\begin{aligned}
    &W_2^2(\mathcal{N}(\boldsymbol{\mu}^{i,l},\boldsymbol{\sigma}^{i,l}), \mathcal{N}(\boldsymbol{\mu}^{j,l},\boldsymbol{\sigma}^{j,l}))\\ 
    = &||\boldsymbol{\mu}^{i,l} - \boldsymbol{\mu}^{j,l}||^2 +\\
    & tr(\boldsymbol{\sigma}^{i,l} + \boldsymbol{\sigma}^{j,l} - 2((\boldsymbol{\sigma}^{i,l})^{1/2}\boldsymbol{\sigma}^{j,l}(\boldsymbol{\sigma}^{i,l})^{1/2} )^{1/2}),
\end{aligned}
    \label{eqa:W}
\end{equation}  
where $tr$ is the trace of the matrix. Obviously, it is too difficulty to perform efficient calculations. Similar to BN, we perform approximations and consider that each channel is independent of each other. Therefore, $\boldsymbol{\sigma}^{i,l}$ is a diagonal matrix, i.e. $\boldsymbol{\sigma}^{i,l} = Diag(\mathbf{r}^{i,l})$. Now, we have
\begin{equation}
\begin{aligned}
    &W_2^2(\mathcal{N}(\boldsymbol{\mu}^{i,l},\boldsymbol{\sigma}^{i,l}), \mathcal{N}(\boldsymbol{\mu}^{j,l},\boldsymbol{\sigma}^{j,l}))\\ 
    = &||\boldsymbol{\mu}^{i,l} - \boldsymbol{\mu}^{j,l}||^2 +
    ||\sqrt{\mathbf{r}^{i,l}} - \sqrt{\mathbf{r}^{j,l}}||_2^2.
\end{aligned}
    \label{eqa:Wapprox}
\end{equation}  

Therefore, we have
\begin{equation}
\begin{aligned}
    d_{i,j}&= \sum_{l=1}^L W_2(\mathcal{N}(\boldsymbol{\mu}^{i,l},\boldsymbol{\sigma}^{i,l}), \mathcal{N}(\boldsymbol{\mu}^{j,l},\boldsymbol{\sigma}^{j,l}))\\ 
    &= \sum_{l=1}^L (||\boldsymbol{\mu}^{i,l} - \boldsymbol{\mu}^{j,l}||^2 +
    ||\sqrt{\mathbf{r}^{i,l}} - \sqrt{\mathbf{r}^{j,l}}||_2^2)^{1/2}.
\end{aligned}
    \label{eqa:dij}
\end{equation}

Big $d_{i,j}$ means the distribution distance between the $i$th client and the $j$th client is large. Therefore, the bigger $d_{i,j}$ is, the less similar two clients are, which means the smaller $w_{i,j}$ is. And we set $\tilde{w}_{i,j}$ as the inverse of $d_{i,j}$, i.e. $\tilde{w}_{i,j} = 1 / d_{i,j}$. Since, $d_{i,i} = 0$, we let $w_{i,i} = \lambda$ and $\sum_{j=1,j\neq i}^N w_{i,j} = 1 - \lambda$. Normalize $\tilde{w}_i$ and we have 
\begin{equation}
w_{i,j}=
\begin{cases}
\lambda,&j=i\\
(1 - \lambda) \times \frac{\tilde{w}_{i,j}}{\sum_{j=1,j\neq i}^N \tilde{w}_{i,j} }, & j\neq i
\end{cases}
\label{eqa:weight}
\end{equation}

We denote this weighting method as \method. Similarly, for $\mathbf{z}^i$, we can obtain the corresponding $\mathbf{W}$ and we denote this method as d-\method.

\subsection{Evaluate Weights Without a Pretrained Model}
Sometimes, there does not exist a pretrained model. In this situation, we can evaluate weights with models trained from several rounds of FedBN.

\begin{figure}[htbp!]
	\centering
	\includegraphics[width=0.35\textwidth]{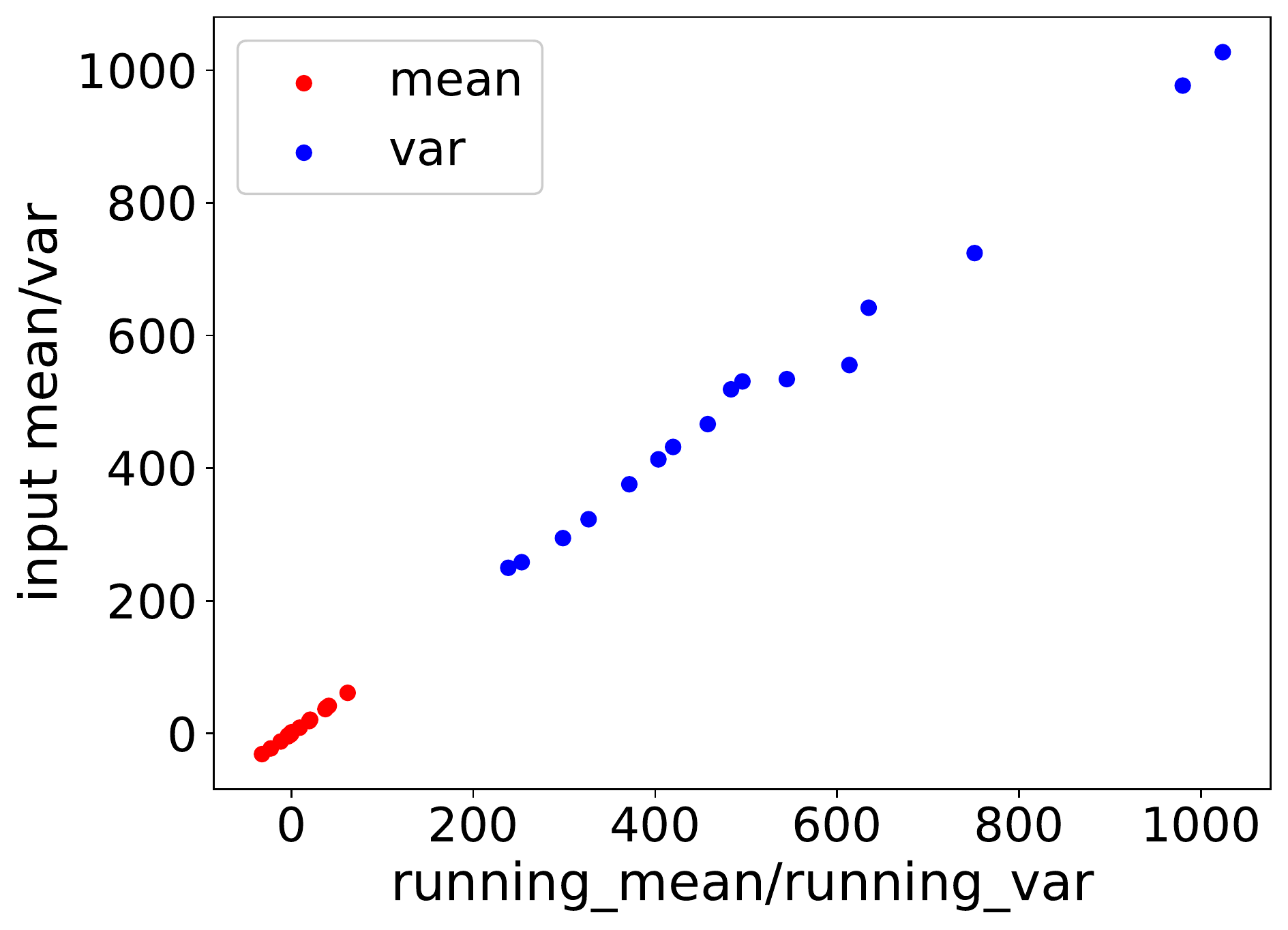}
	\caption{The running mean, running var of a BN layer and the inputs statistics of the corresponding layer in a client model.}
	\label{fig:fedbnweight}
\end{figure}

As we can see from \figurename~\ref{fig:fedbnweight}, the running mean of the BN layer has a positive correlation with the statistical mean of the corresponding layer's inputs. And the variance has a similar relationship. From this, we can use running means and running variances of the BN layers instead of the statistics respectively. Therefore, we can perform several rounds of FedBN~\cite{li2021fedbn} which preserves local batch normalization, and utilize parameters of BN layers replacing the statistics when there does not exist a pretrained model. We denote this extension as f-\method.

\section{Experiments}
\subsection{Datasets}
\begin{figure}[ht!]
	\centering
	\subfigure[PAMAP]{
		\label{fig:pamapdata}
		\includegraphics[width=0.22\textwidth]{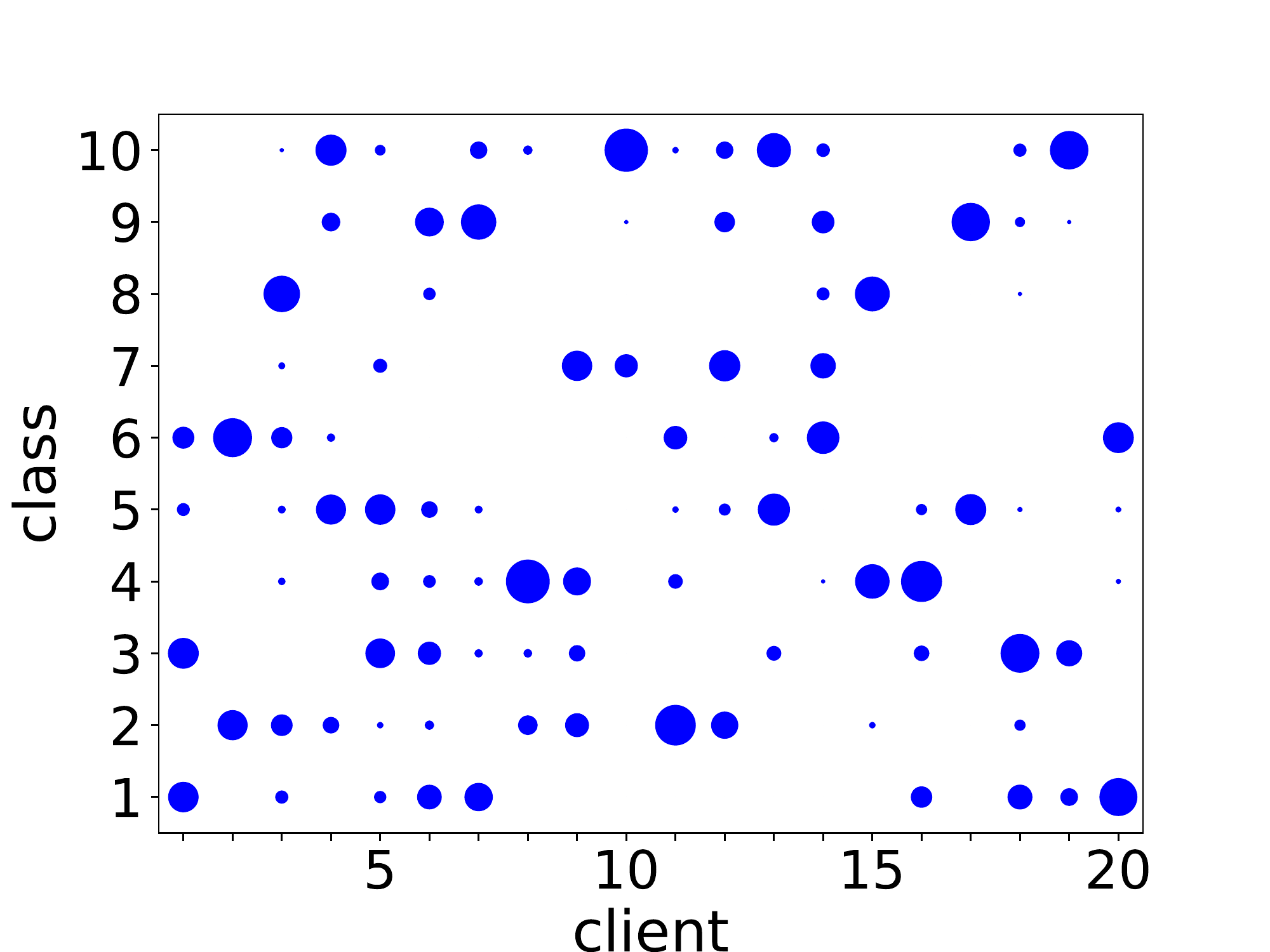}
	}
	\subfigure[COVID-19]{
		\label{fig:coviddata}
		\includegraphics[width=0.22\textwidth]{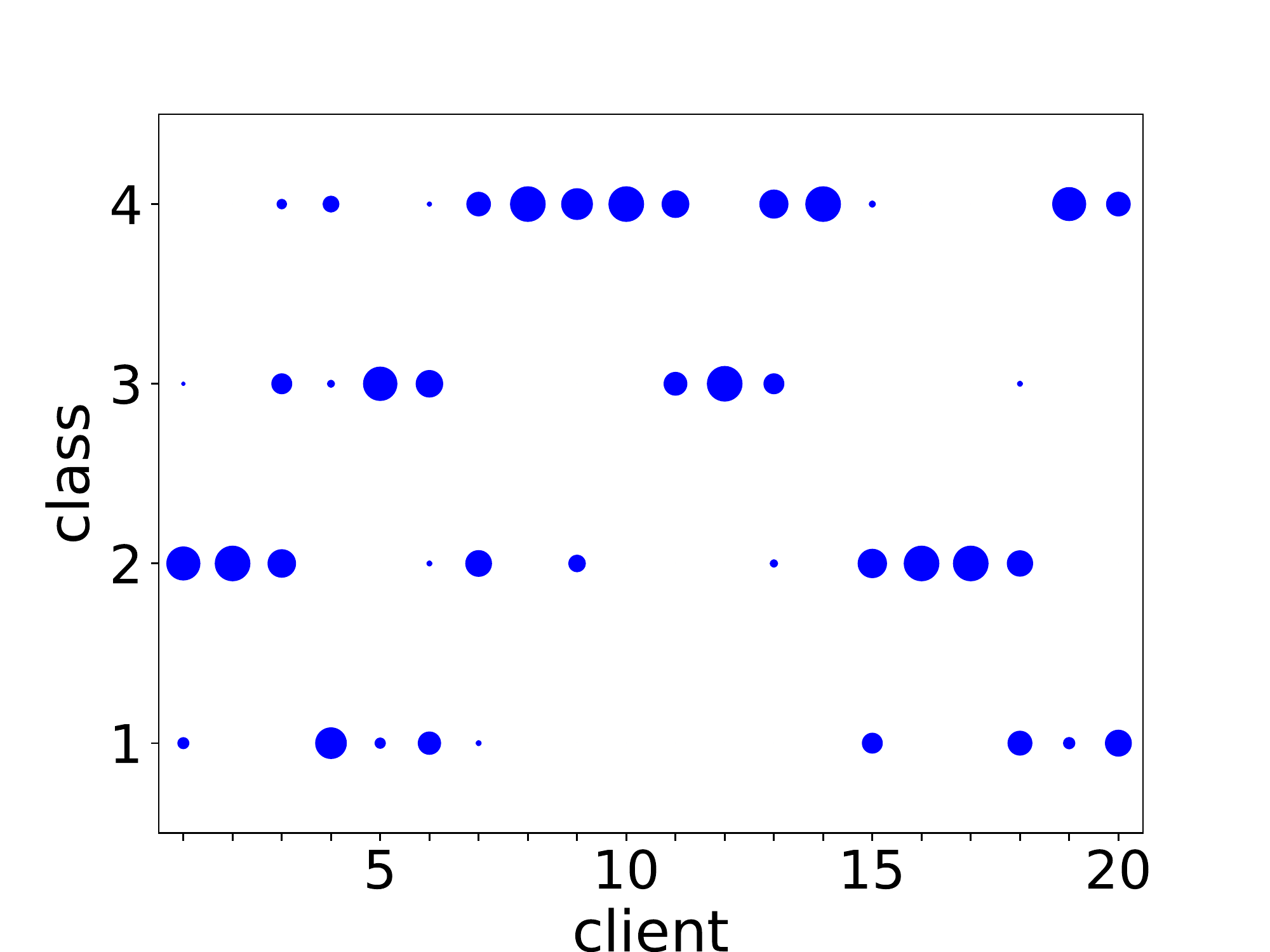}
	}
	\vspace{-.2in}
	\caption{The number of samples per class allocated to each client (indicated by dot size).}
	\label{fig:datasplit}
\end{figure} 
\paragraph{PAMAP.} We adopt a public human activity recognition dataset called PAMAP~\cite{reiss2012introducing}. The PAMAP dataset contains data of 18 different physical activities, performed by 9 subjects wearing 3 inertial measurement units and a heart rate monitor. We use data of 3 inertial measurement units which are collected at a constant rate of 100Hz to form data containing 27 channels. We exploit the sliding window technique and filter out 10 classes of data to obtain 17639 instances in total. In order to construct the problem situation in \method, we use the Dirichlet distribution as in~\cite{yurochkin2019bayesian}  to create disjoint non-iid. client training data. \figurename~\ref{fig:pamapdata} visualizes how samples are distributed among 20 clients for PAMAP. In each client, half of the data are used to train and the remaining data are for testing.
\paragraph{COVID-19.}
We also adopt a public COVID-19 posterior-anterior chest radiography images dataset~\cite{2020Curated}. This is a combined curated dataset of COVID-19 Chest X-ray images obtained by collating 15 public datasets and it contains 9,208 instances of four classes (1281 COVID-19 X-Rays, 3270 Normal X-Rays, 1656 viral-pneumonia X-Rays, and 3001 bacterial-pneumonia X-Rays) in total. In order to construct the problem situation in \method, we split the dataset similar to PAMAP. \figurename~\ref{fig:coviddata} visualizes how samples are distributed among 20 clients for COVID-19. In each client, half of the data are used to train and the remaining data are for testing.
\begin{table*}[!t]
\centering
\resizebox{\textwidth}{!}{
\begin{tabular}{lrrrrrrrrrrrrrrrrrrrr|r}
\toprule
Client& 1&2&3&4&5&6&7&8&9&10&11&12&13&14&15&16&17&18&19&20&avg\\
\midrule
Base&\textbf{92.86}&17.68&\textbf{100.00}&\textbf{83.52}&18.78&77.66&\textbf{95.05}&17.58&\textbf{92.39}&\textbf{93.37}&29.12&\textbf{84.78}&\textbf{98.90}&24.18&\textbf{98.91}&\textbf{98.90}&41.44&\textbf{93.62}&\textbf{85.71}&37.02&69.07\\
FedAvg&60.27&62.36&50.56&73.98&74.27&62.90&64.03&87.78&74.49&64.71&65.24&63.35&68.33&64.79&63.12&85.26&66.21&59.64&67.87&72.46&67.58\\
FedBN&60.72&62.59&50.34&73.53&74.72&62.44&62.90&88.24&74.27&64.48&65.69&62.90&68.33&65.24&62.44&85.94&65.99&59.64&68.10&72.69&67.56\\
FedProx&60.50&62.36&50.34&73.98&73.81&61.76&63.57&87.78&74.27&64.71&66.37&63.12&68.33&65.69&62.44&85.49&66.21&59.41&67.87&72.46&67.52\\
FedPer&48.31&\textbf{97.51}&61.40&47.29&58.47&23.98&49.55&91.86&51.24&77.60&\underline{89.16}&57.92&42.53&49.44&58.60&86.62&77.32&52.38&73.08&\textbf{97.52}&64.59
\\
\midrule
\method&\underline{77.20}&77.55&77.43&79.64&\textbf{81.94}&\underline{79.86}&\underline{86.20}&\textbf{95.02}&85.33&69.23&\textbf{91.42}&79.41&74.43&\underline{69.75}&81.67&\underline{94.10}&\textbf{82.77}&\underline{75.74}&77.15&86.91&\textbf{81.14}\\
d-\method&64.33&77.55&\underline{78.33}&77.38&\underline{79.91}&\textbf{80.77}&85.52&92.53&\underline{86.23}&69.23&87.58&\underline{80.09}&\underline{74.66}&\textbf{70.43}&\underline{83.71}&93.88&\underline{80.50}&74.60&\underline{78.73}&87.13&\underline{80.16}\\
f-\method&64.11&\underline{77.78}&69.53&\underline{79.86}&77.88&74.43&84.62&\underline{93.67}&74.04&\underline{81.00}&79.91&71.95&74.21&62.98&78.05&89.57&79.59&68.71&71.95&\underline{87.81}&77.08\\
\bottomrule
\end{tabular}
}
\caption{Activity recognition results of 20 clients on PAMAP. Bold means the best result while underline means the second best result.}
\label{tab:pamapresults}
\end{table*}
\subsection{Implementations Details and Comparison Methods}
For PAMAP, we adopt a CNN for training and predicting. The network is composed of two convolutional layers, two pooling layers, two batch normalization layers, and two fully connected layers. For COVID-19, we adopt Alexnet~\cite{krizhevsky2012imagenet}. We use a three-layer fully connected neural network as the classifier with two BN layers after the first two fully connected layers following~\cite{li2021fedbn}. For model training, we use the cross-entropy loss and SGD optimizer with a learning rate of $10^{-2}$. If not specified, our default setting for local update epochs is $E = 1$ where $E$ means training epochs in one round. In addition, we randomly select $20\%$ of the data to train a model of the same architecture as the pretrained model.

We did not compare with FedHealth~\cite{chen2020fedhealth} since our method focuses on more general setting where clients do not share large volume of datasets. We compare three extensions of our method with five methods including common federated learning methods and some federated learning methods designed for non-iid data particularly:
\begin{itemize}
    \item Base: Each client uses local data to train local models.
    \item FedAvg~\cite{mcmahan2017communication}: The cloud aggregate all client models without any particular operations for non-iid data.
    \item FedBN~\cite{li2021fedbn}: Each client preserves the local batch normalization.
    \item FedProx~\cite{li2018federated}: Allow partial information aggregation and add a proximal term to FedAvg.
    \item FedPer~\cite{arivazhagan2019federated}: Each client preserves some local layers.
\end{itemize}

\subsection{Classification Accuracy}

The classification results for each client on PAMAP are shown in \tablename~\ref{tab:pamapresults}. From these results, we have the following observations: 1) No matter how the weights are calculated, our method achieves the best effects on average. It is obvious that our method significantly outperforms other methods with a remarkable improvement (over $\mathbf{10}\%$ on average). 2) In some clients, the base method achieves the best test accuracy. As it can be seen from \figurename~\ref{fig:pamapdata}, the distributions on the clients are very inconsistent, which inevitably leads to generate the different difficulty levels in different clients. And some distributions in the corresponding clients are so easy that only utilizing the local data can achieve the ideal effects. 3) FedBN does not achieve the desired results. This could be caused by that FedBN is designed for the feature shifts while our experiments are mainly set in the label shifts. 
\begin{table}[!htbp]
\centering
\resizebox{0.4\textwidth}{!}{
\begin{tabular}{lllll}
\toprule
Method&Base&FedAvg&FedBN&FedProx\\
\hline
avg&92.70&86.48&82.26&86.15\\
\midrule \midrule
Method&FedPer&\method&d-\method&f-\method\\
\hline
avg&88.51&\underline{94.82}&\textbf{94.97}&93.5\\
\bottomrule
\end{tabular}}
\caption{Average accuracy of 20 clients on COVID-19}
\label{tab:covidresults}
\end{table}

The classification results for each client on COVID-19 are shown in \tablename~\ref{tab:covidresults}. From these results, we have the following observations: 1) No matter how the weights are calculated, our method achieves the best effects. However, in those experiments, our methods only have slight improvements. The classification accuracy only improves by $2\%$ on average. It may be caused by that the COVID-19 dataset has much fewer classes compared with PAMAP. Henceforth, the task is much easier. 2) The base method achieves acceptable results in this dataset. From \figurename~\ref{fig:coviddata}, we can see that there are only a few classes in some clients, which means it is easy to obtain the desired results with few data. 3) FedBN gets the worst results. This demonstrates that FedBN is not good at dealing with label shifts. And FedBN does not consider the similarities among different clients. 

\subsection{Ablation Study}
\begin{figure}[ht!]
	\centering
	\subfigure[Average Accuracy]{
		\label{fig:weffa}
		\includegraphics[height=0.15\textwidth]{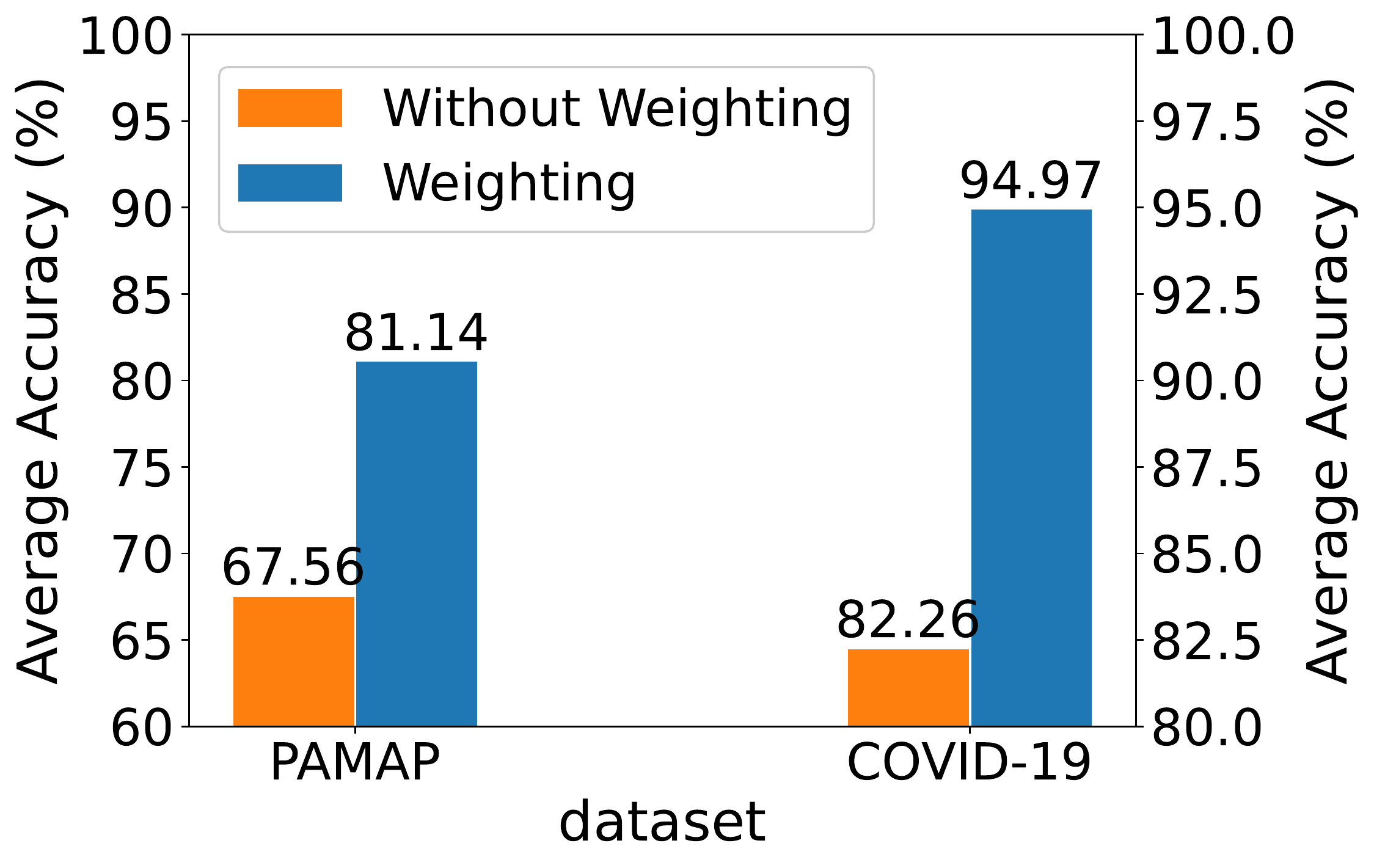}
	}
	\subfigure[Client Acc on PAMAP]{
		\label{fig:weff}
		\includegraphics[height=0.15\textwidth]{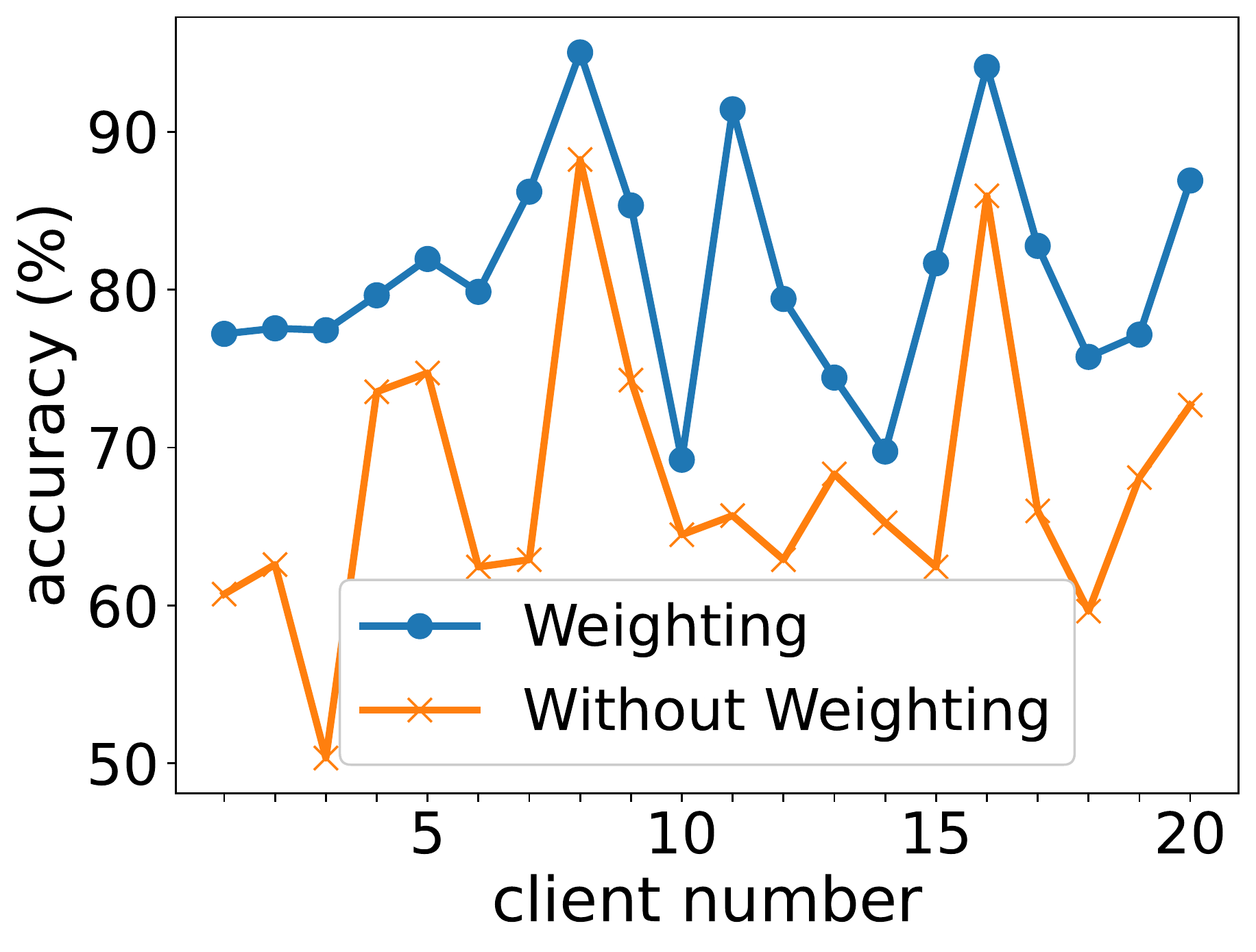}
	}
	\vspace{-.2in}
	\caption{The effects of weighting.}
	\label{fig:weightingeff}
\end{figure}

\paragraph{Effects of Weighting.}
To demonstrate the effect of weighting which considers the similarities among the different clients, we compare the average accuracy on PAMAP and COVID-19 between the experiments with it and without it. Without weighting, our method degenerates to FedBN. From \figurename~\ref{fig:weffa}, we can see that our method performs much better than FedBN which does not include the weighting part. Moreover, from \figurename~\ref{fig:weff}, we can see our method performs better than FedBN on all clients. These results demonstrate that our method with weighting can cope with the label shifts while FedBN cannot deal with this situation, which means our method  is more applicable and effective. 
\vspace{-.1in}
\begin{figure}[ht!]
	\centering
	\subfigure[Average Accuracy]{
		\label{fig:bneffa}
		\includegraphics[height=0.15\textwidth]{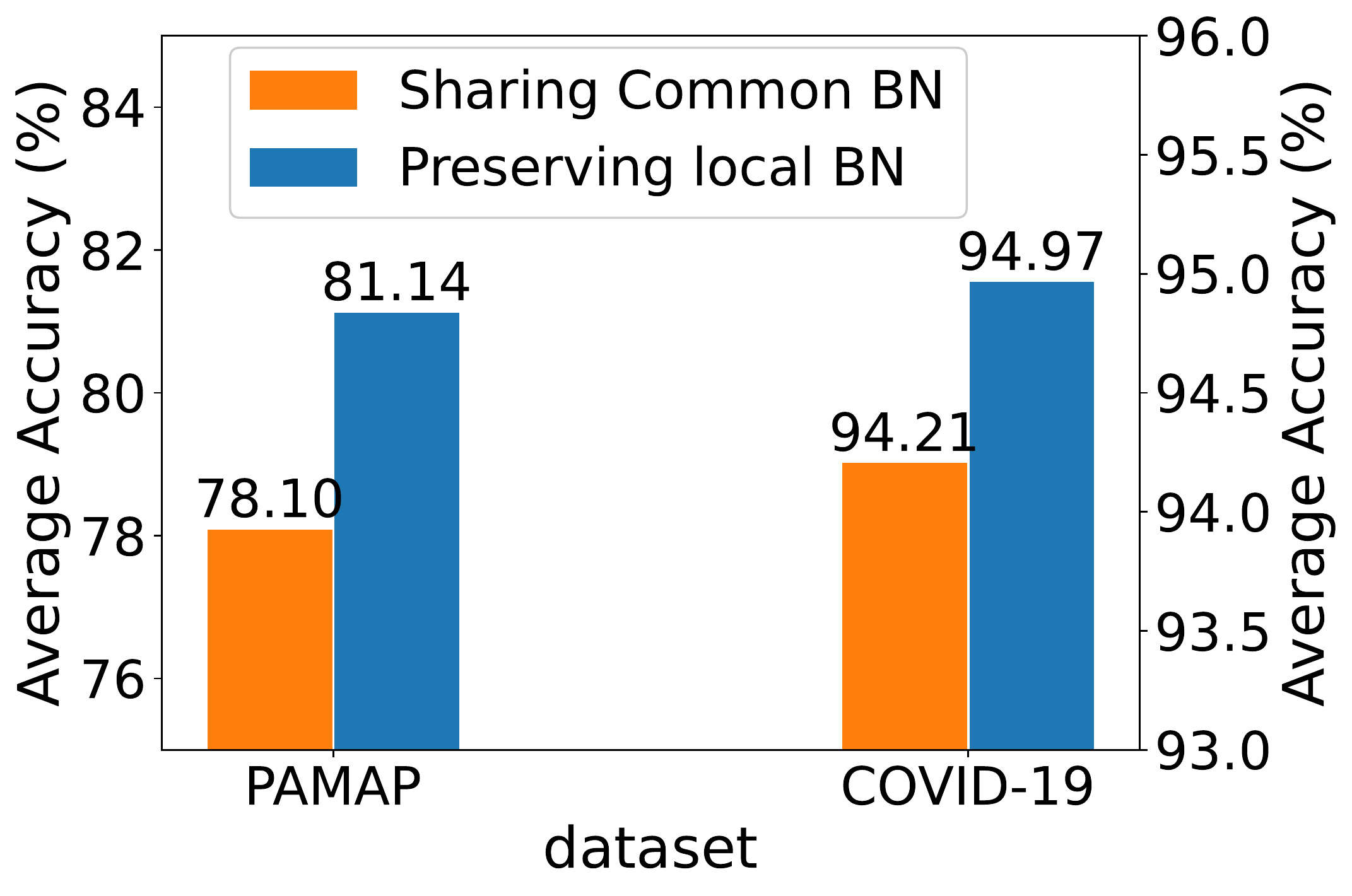}
	}
	\subfigure[Client Acc on PAMAP]{
		\label{fig:bneff}
		\includegraphics[height=0.15\textwidth]{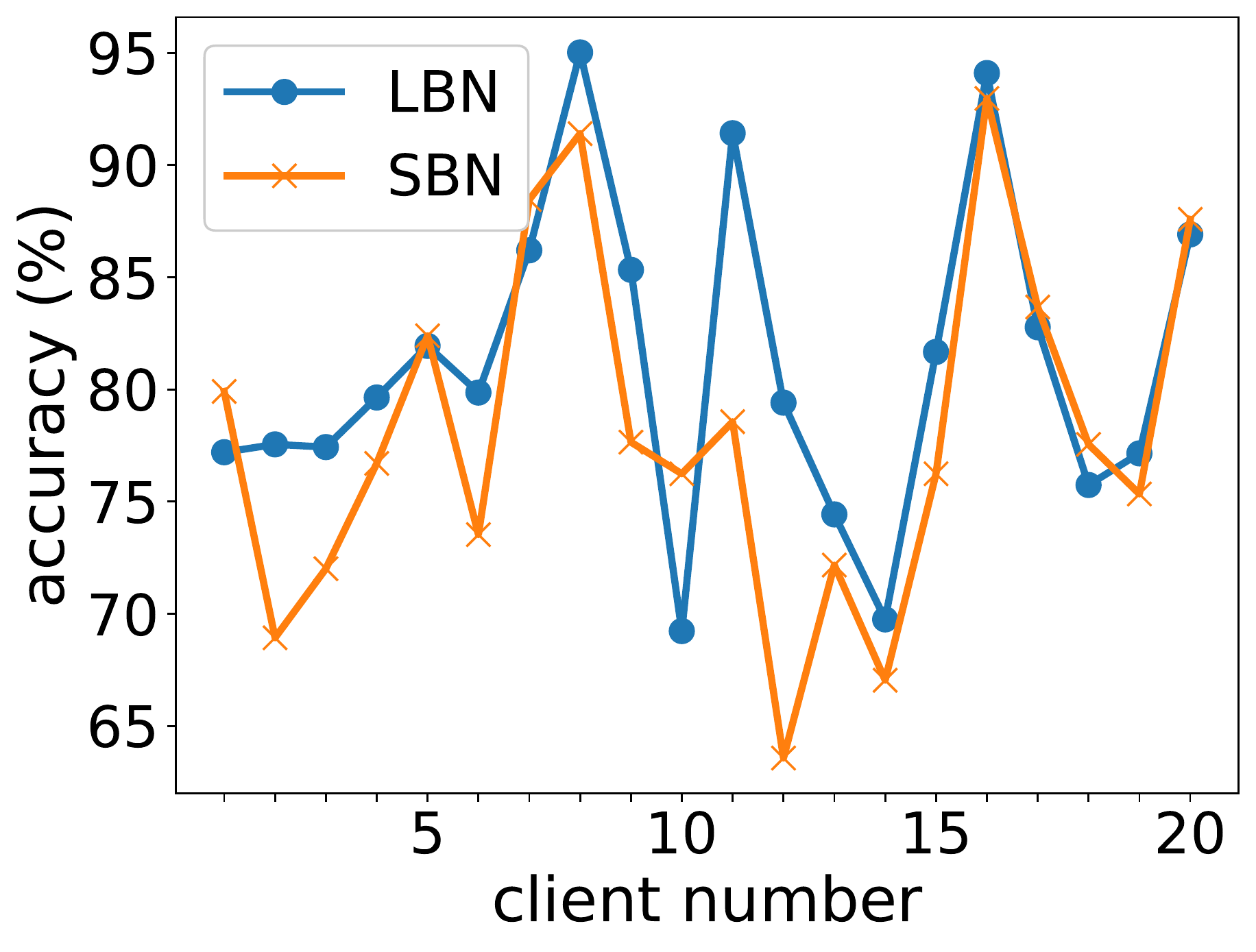}
	}
	\vspace{-.2in}
	\caption{The effects of preserving local batch normalization.}
	\label{fig:BNabeff}
\end{figure} 
\paragraph{Effects of Preserving Local Batch Normalization.}
In this part, we illustrate the function of preserving local batch normalization. \figurename~\ref{fig:bneffa} shows the average accuracy between the experiments with preserving local batch normalization and the experiments with sharing common batch normalization while \figurename~\ref{fig:bneff} shows the results on each client. LBN means preserving local batch normalization while SBN means sharing common batch normalization. Obviously, the improvements are not particularly significant compared with weighting. This may be caused by there mainly exist the label shifts in our experiments while preserving local batch normalization is for the feature shifts. However, our method still has a slight improvement, which shows the superiority of our method.

\subsection{Parameter Sensitivity}
\begin{figure}[ht!]
	\centering
	\subfigure[Local Epochs]{
		\label{fig:s-lepc}
		\includegraphics[width=0.22\textwidth]{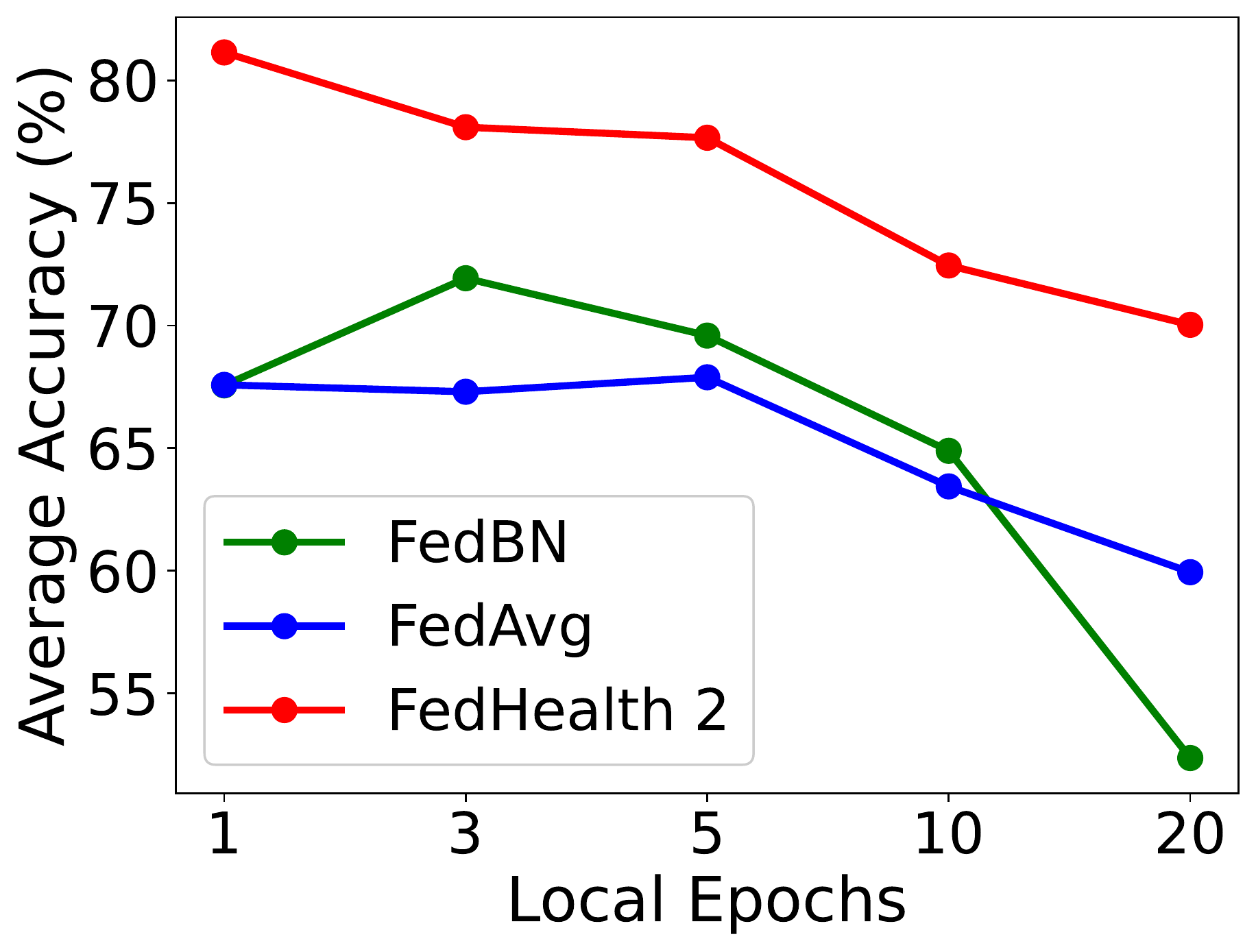}
	}
	\subfigure[Client Number]{
		\label{fig:s-clients}
		\includegraphics[width=0.22\textwidth]{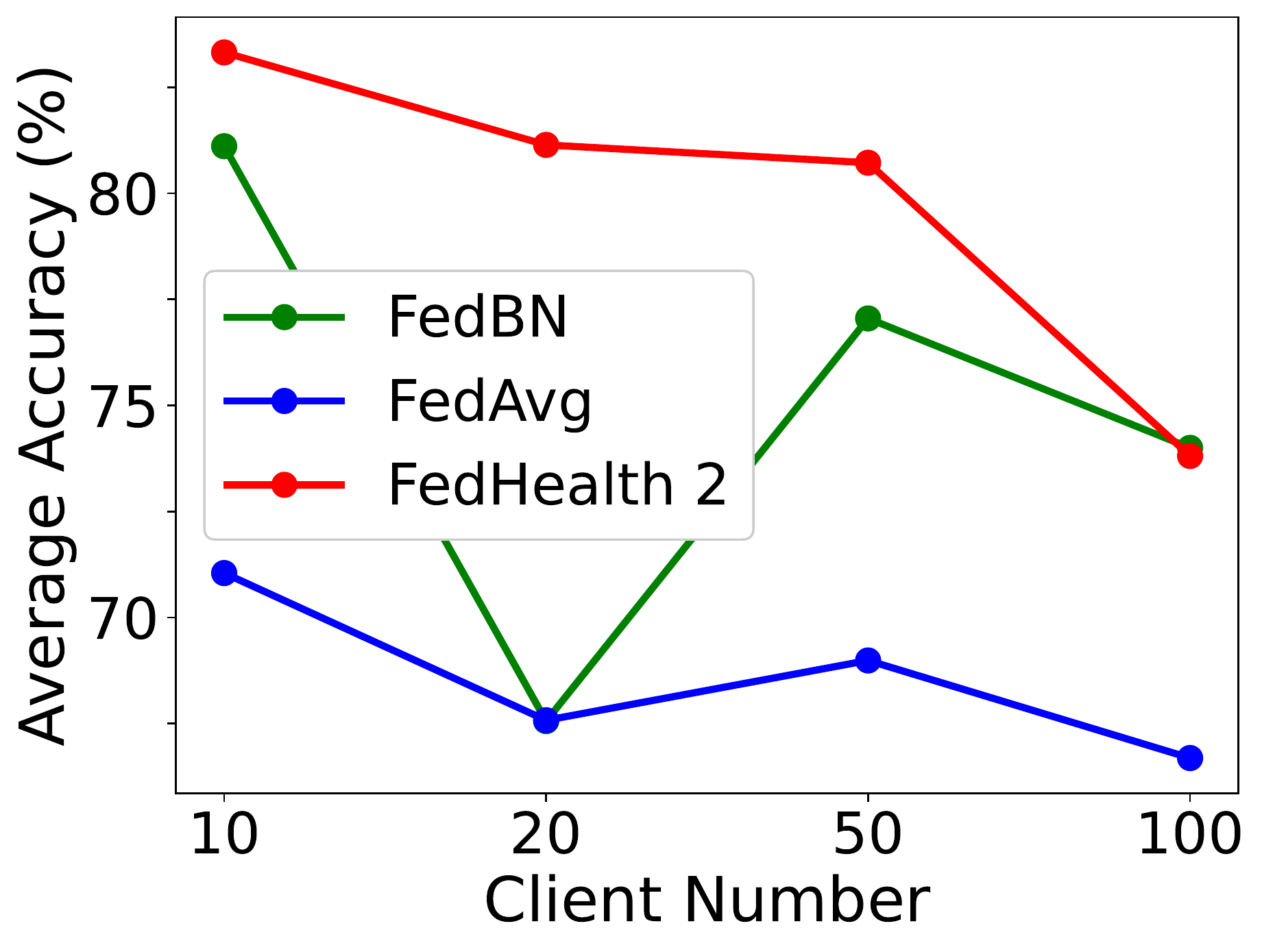}
	}
	\subfigure[$\lambda$]{
		\label{fig:s-lambda}
		\includegraphics[width=0.22\textwidth]{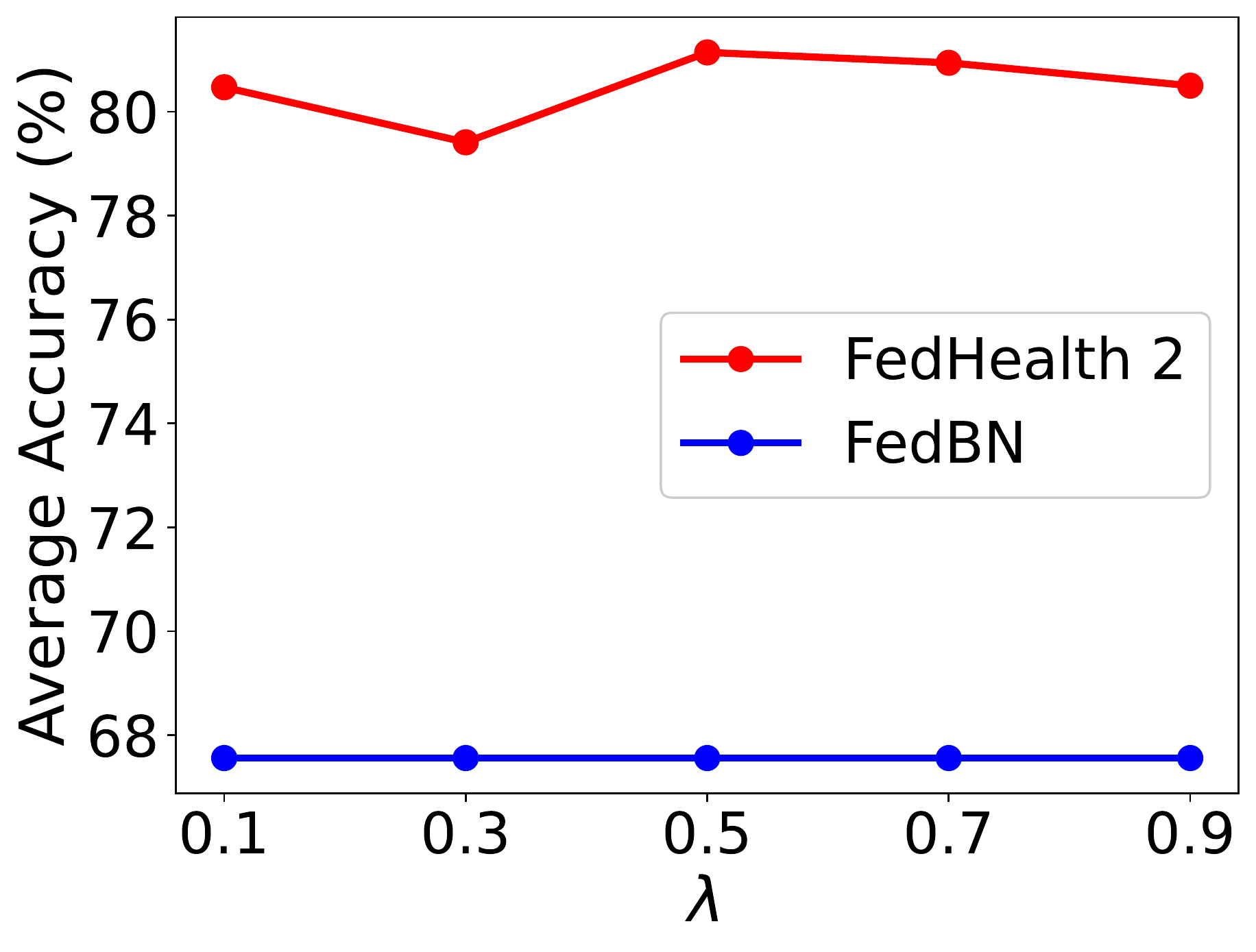}
	}
	\subfigure[$\lambda$]{
		\label{fig:s-lambdafun}
		\includegraphics[width=0.22\textwidth]{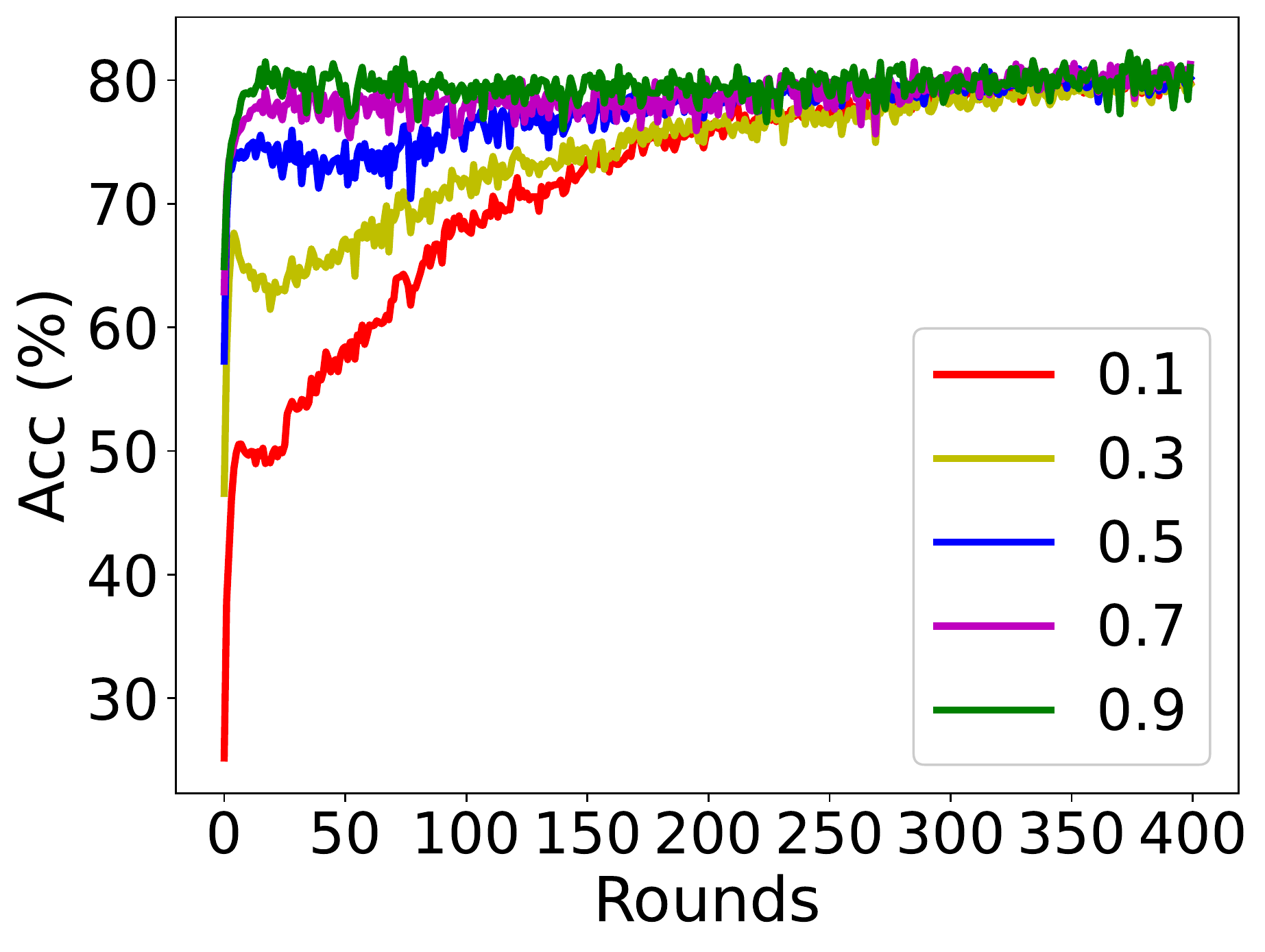}
	}
	\vspace{-.2in}
	\caption{Influence of different hyper-parameters.}
	\label{fig:sensity}
\end{figure} 
In this section, we evaluate the parameter sensitivity of \method and we use \method. Our method is affected by three parameters: local epochs, client number, and $\lambda$. We change one parameter and fix the other parameters. In \figurename~\ref{fig:s-lepc}, we can see our method is best and it is descending with local epochs increasing, which may be caused that we keep the total number of the epochs unchanged and the communication among the clients are insufficient. From \figurename~\ref{fig:s-clients}, we can see that our method still achieves acceptable results. When the client numbers increase, our method goes down which may due to that few data in local clients make the weight estimation inaccurate. And we may take f-\method instead. \figurename~\ref{fig:s-lambda}-\ref{fig:s-lambdafun} demonstrates $\lambda$ slightly affects the average accuracy of our method while it can change the convergence rate. The results reveal that 
\method is more effective and robust than other methods under different parameters in most cases.
\subsection{Convergence}

\begin{figure}[htbp!]
	\centering
	\includegraphics[width=0.3\textwidth]{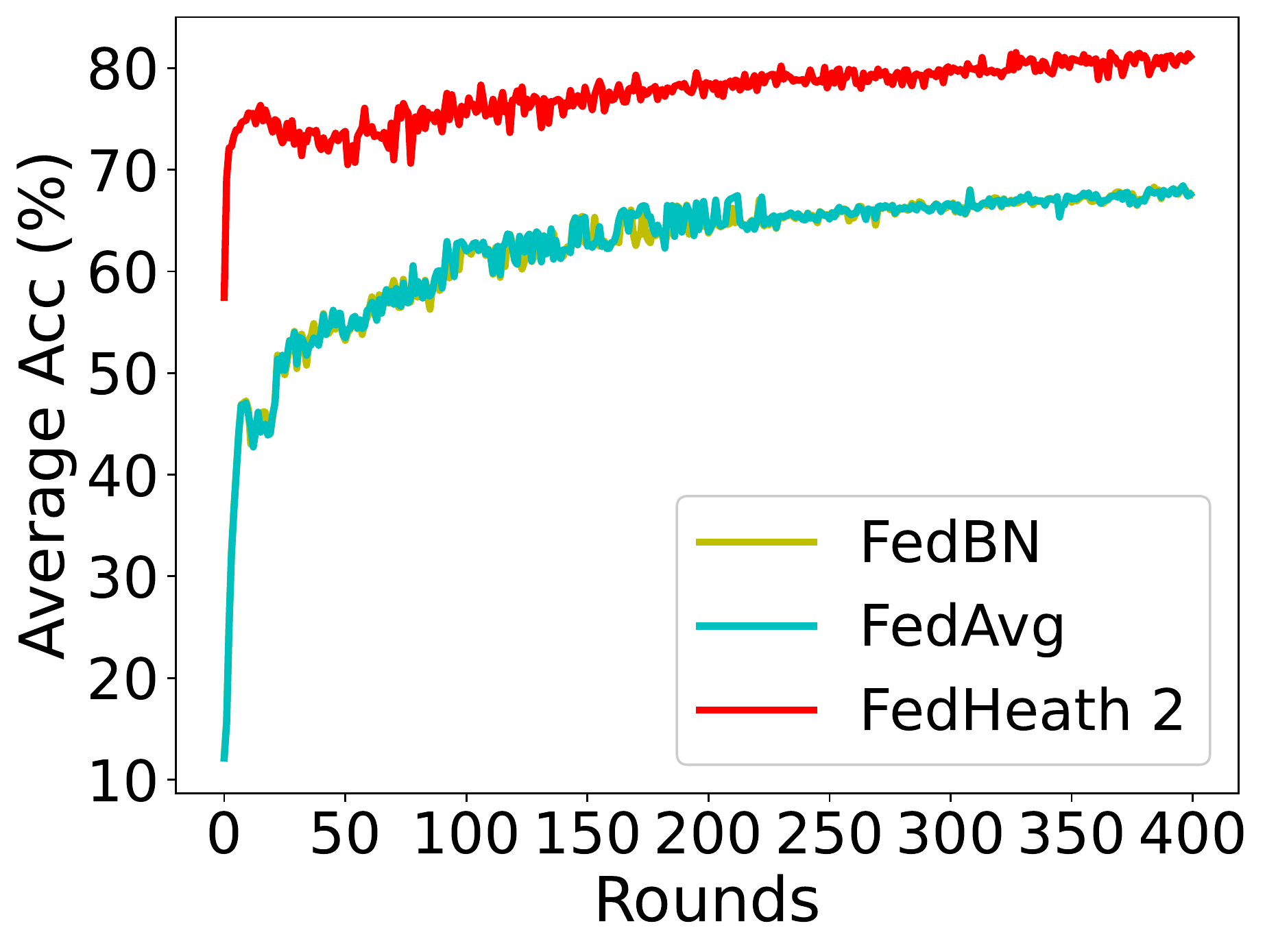}
	\vspace{-.05in}
	\caption{Convergence.}
	\label{fig:converge}
\end{figure}

In this section, we investigate the convergence. In \figurename~\ref{fig:converge}, we can see our method almost convergences in the $10$th round. And in the actual experiments, 20 rounds are enough for our method while FedBN needs over 400 rounds.

\section{Conclusions And Future Work}

In this article, we propose \method, a weighted federated transfer learning algorithm via batch normalization for personalized healthcare. \method aggregates the data
from different organizations without compromising privacy security and achieves relatively personalized model learning through combing considering similarities and preserving local batch normalization. Experiments have evaluated the effectiveness of \method.
\method follows FedHealth and continues to move forward in the direction of federated learning for healthcare. In the future, we plan to apply \method to more personalized and flexible healthcare.

\bibliographystyle{named}
\bibliography{ijcai21}

\end{document}